\documentclass[11pt]{article}

\usepackage{acl}

\usepackage{times}
\usepackage{latexsym}
\usepackage[T1]{fontenc}
\usepackage[utf8]{inputenc}
\usepackage{microtype}
\usepackage{graphicx}

\usepackage{booktabs}
\usepackage{makecell}
\usepackage{multirow}
\usepackage{amsmath}
\usepackage{xcolor}
\usepackage{enumitem}
\usepackage{tikz}
\usetikzlibrary{arrows.meta, positioning, shapes.geometric, fit, calc, backgrounds}

\newcommand{\wer}{\textsc{WER}}
\newcommand{\ewer}{\textsc{E-WER}}

\newcommand{\react}{\textsc{ReAct}}

\newcommand{\sv}[3]{\makecell{#1\\[-3pt]{\scriptsize(#2\,/\,#3)}}}
\newcommand{\svs}[2]{\makecell{#1\\[-3pt]{\scriptsize(#2)}}}

\title{RECOVER: Robust Entity Correction via agentic Orchestration of\\ hypothesis Variants for Evidence-based Recovery}

\author{Abhishek Kumar \quad Aashraya Sachdeva \\
  Observe.AI, India \\
  \texttt{\{abhishek.kumar, aashraya.sachdeva\}@observe.ai}}

\begin{document}
\maketitle

\begin{abstract}
Entity recognition in Automatic Speech Recognition (ASR) is challenging for rare and domain-specific terms. In domains such as finance, medicine, and air-traffic control, these errors are costly. If an entity is entirely absent from the ASR output, text-only post-correction has little signal to recover it. To address this, we introduce RECOVER, an agentic post-ASR entity-correction framework. It leverages multiple hypotheses as evidence from ASR, retrieves relevant entities, proposes Large Language Model (LLM) correction under constraints, filters them through deterministic guardrails and uses an LLM critic that judges each surviving edit and can trigger a targeted feedback round to recover an entity the first pass missed. The hypotheses are used by one of several strategies, namely 1-Best, Entity-Aware Select, Recognizer Output Voting Error Reduction (ROVER) Ensemble, LLM-Select and Adaptive Routing. Evaluated across five diverse datasets over a Whisper-large-v3 baseline, it achieves up to 49\% relative reduction in entity-phrase word error rate (E-WER) and increases entity recall by up to 14 percentage points while preserving the overall word error rate (\wer), with LLM-Select the strongest strategy.
\end{abstract}

\section{Introduction}
\label{sec:intro}

Modern end-to-end ASR systems such as Whisper~\cite{radford2023robust}, wav2vec~2.0~\cite{baevski2020wav2vec}, and Conformer-based models~\cite{gulati2020conformer, burchi2021efficient} achieve impressive \wer{} on general read- and conversational-speech benchmarks such as LibriSpeech~\cite{panayotov2015librispeech} and Common Voice~\cite{ardila2020common}. However, their performance degrades on entity phrases, i.e.\ proper nouns, brand names, and domain-specific terminology that are rare or absent from training data~\cite{sainath2019two, del2021earnings}. In domains like finance, air-traffic control (ATC), and medicine, entity errors are costly and often appear as small but critical near-misses (e.g.\ \textit{cytiba}$\rightarrow$\textit{cytiva}, \textit{linear zolid}$\rightarrow$\textit{linezolid}).

One line of work addresses these errors at decode time through contextual biasing~\cite{jain2020contextual,le2021contextualized,fox2022improving,huang2023contextualized}, but such methods require access to the recognizer's internals and are unavailable when the ASR is a black-box or production service (Section~\ref{sec:related}).

Post-ASR correction with LLMs~\cite{yang2023generative,ma2025asr,chen2023hyporadise} is an attractive black-box alternative, but it faces a fundamental evidence limitation, that is, if an entity is deleted or heavily corrupted in the 1-best output, a text-only corrector has no signal to recover it. One approach is to expose the corrector to several ASR hypotheses, since an entity missed in one may survive in another. We obtain this diversity from a single black-box ASR via temperature sampling, without an ensemble or decoder access.

A second issue is reliability, as LLM correction is prone to hallucination and over-correction unless tightly constrained~\cite{pu2023multi,manakul2023selfcheckgpt}, and a single forward pass cannot recover when its own output is wrong. We therefore constrain edits to a verified entity vocabulary and add a verification stage. Agents that use tools and act on their observations have proven effective~\cite{yao2022react,schick2023toolformer}, but their use in speech has so far been limited to conversational tasks rather than post-ASR entity recovery (Section~\ref{sec:related}).

These observations motivate RECOVER, a post-ASR entity-correction framework that leverages multiple hypotheses from an ASR system when available, retrieves candidate entities for each segment, fuses the hypotheses into a working transcript, proposes constrained edits, and verifies them with deterministic guardrails and an LLM critic. Our contributions are:
\begin{itemize}[leftmargin=*, itemsep=2pt, topsep=2pt]
\item \textbf{Multi-hypothesis evidence.} We show that multiple hypotheses decoded from the same audio carry complementary entity errors that can be exploited for correction. We obtain them via temperature sampling of a single ASR model, but the framework is agnostic to how hypotheses are produced.
\vspace{-1mm}

\item \textbf{Agentic controller.} A controller composes a tool suite including entity retrieval, multi-hypothesis fusion, constrained proposal, deterministic verification, and an LLM critic. It runs a bounded feedback round when the critic flags a recoverable entity the first pass missed, and under adaptive routing it lets an LLM choose the fusion strategy per segment.
\vspace{-1mm}

\item \textbf{Fusion strategies.} 1-Best, Entity-Aware Select, ROVER Ensemble, and LLM-Select differ in how hypotheses are combined (deterministic selection, token-level merge, or LLM selection). For Adaptive Routing, an LLM picks one of these strategies per segment using the hypotheses and retrieved candidates.
\vspace{-1mm}

\end{itemize}

We conduct comprehensive evaluations across five diverse datasets spanning financial, ATC, medical, general speech, and dialogue domains, reporting up to 49\% relative E-WER reduction under temperature sampling with Whisper-large-v3. We also provide analyses of the feedback, guardrails, critic, the retrieval hyperparameters, comparison with external baselines, the effect of LLM and ASR model size, cost and latency.

\vspace{-1mm}

\section{Related Work}
\label{sec:related}
\vspace{-2mm}

\paragraph{Contextual biasing in ASR.}
Decode-time biasing improves rare-word recognition through deep biasing in transducers~\cite{jain2020contextual,huang2023contextualized}, alternate-spelling prediction~\cite{fox2022improving}, and trie-based shallow fusion~\cite{le2021contextualized,thorbecke2025unifying}. More recent work matches biasing phrases directly against the decoder's hypothesis prefix~\cite{wang2024contextbiasing}, and folds entity recognition into the recognizer itself~\cite{ayache2024whispener}. All require white-box access to the recognizer. RECOVER instead operates on the text output of a black-box ASR.

\vspace{-1mm}
\paragraph{LLM-based post-ASR correction.}
Generative correction and N-best reranking with LLMs have become strong post-processing tools~\cite{ma2025asr,yang2023generative,pu2023multi,chen2023hyporadise,xu2025large,azmi2025llm}. We adopt this paradigm but constrain edits to a verified entity vocabulary to suppress hallucination, and add a verification stage before an edit is committed.
\vspace{-6mm}
\paragraph{Agents and tool use.}
\react~\cite{yao2022react} combines reasoning and acting, Toolformer~\cite{schick2023toolformer} learns when to call tools, and Reflexion~\cite{shinn2023reflexion} adds verbal self-feedback~\cite{wang2024survey,derouiche2025agentic}. RECOVER borrows tool use and acting on verifier feedback from this line of work and applies them to post-ASR entity correction.

\vspace{-1mm}

\section{Method}
\label{sec:approach}
\vspace{-2mm}

ASR systems often misrecognize entity phrases and rare or out-of-vocabulary (OOV) tokens. Given a list of known entity phrases and one or more ASR hypotheses, our goal is to correct entity-level errors while leaving the rest of the text intact. To address this, we propose a framework, RECOVER, as in Figure~\ref{fig:agentic} which consists of a {tool suite} (Sections~\ref{sec:multihyp}--\ref{sec:tools}) and a {controller} that composes those tools.
\vspace{-2mm}

\subsection{Multi-Hypothesis Generation}
\label{sec:multihyp}
\vspace{-1mm}

The framework accepts $N$ ASR hypotheses per audio segment. Hypotheses may come from temperature sampling of a single ASR model, diverse beam search, or an ensemble of different ASR systems. The only requirement when $N>1$ is that they contain \emph{complementary} errors, so an entity missed in one may be correct in another. Our main experiments use five hypotheses from temperature sampling of the Whisper model, and Section~\ref{sec:crossasr} takes one from each of five ASR systems.

\begin{figure*}[t]
  \centering
  
  \resizebox{0.92\textwidth}{!}{%
  \begin{tikzpicture}[
    font=\sffamily,
    io/.style={draw, rounded corners=3pt, fill=green!12, minimum height=0.9cm, text width=2.3cm, align=center, line width=0.6pt},
    gen/.style={draw, rounded corners=3pt, fill=yellow!18, minimum height=0.9cm, text width=2.6cm, align=center, line width=0.6pt},
    db/.style={draw, rounded corners=3pt, fill=orange!14, minimum height=0.9cm, text width=2.1cm, align=center, line width=0.6pt},
    ctrl/.style={draw, rounded corners=6pt, fill=blue!12, minimum height=2.9cm, text width=3.5cm, align=center, line width=1.0pt},
    tool/.style={draw, rounded corners=3pt, fill=red!8, minimum height=0.85cm, text width=3.5cm, align=center, line width=0.6pt},
    outb/.style={draw, rounded corners=3pt, fill=cyan!16, minimum height=0.9cm, text width=2.4cm, align=center, line width=0.6pt},
    dec/.style={draw, diamond, aspect=2, fill=gray!12, align=center, inner sep=1pt, line width=0.6pt, text width=1.8cm},
    arr/.style={-{Stealth[length=5pt]}, thick, black!75},
    barr/.style={{Stealth[length=5pt]}-{Stealth[length=5pt]}, thick, black!70},
    lbl/.style={font=\scriptsize\sffamily, text=black!60},
  ]
    \node[io] (audio) {Audio};
    \node[gen, below=0.5cm of audio] (whisper) {Whisper\\$T\!\in\!\{0,.2,.4,.6,.8\}$\\{\scriptsize $N{=}5$ hypotheses}};
    \node[db, below=0.5cm of whisper] (elist) {Entity Phrase List};
    \draw[arr] (audio) -- (whisper);

    \node[ctrl, right=2.4cm of whisper] (controller)
      {\textbf{Controller}\\[5pt]
       \fbox{\parbox{2.9cm}{\centering\scriptsize \textbf{State}\\ working transcript,\\ candidates, accepted /\\ rejected edits, \#rounds}}};
       

    \draw[arr] (whisper.east) -- node[lbl, above]{hypotheses} (controller.west);
    \draw[arr] (elist.east) -- node[lbl, below, pos=0.62]{entities} ([yshift=-0.7cm]controller.west);

    \node[tool, right=2.7cm of controller, yshift=1.55cm] (t1) {\textbf{\textsc{Retrieve}}\\{\scriptsize top-$K$ entity candidates}\\};
    \node[tool, below=0.24cm of t1] (t2) {\textbf{\textsc{Fuse}} \\{\scriptsize 1-Best / Entity-Aware /}\\{\scriptsize ROVER / LLM-Select / Adaptive Routing}};
    \node[tool, below=0.24cm of t2] (t3) {\textbf{\textsc{Propose}}\\{\scriptsize constrained LLM find / replace edits}};
    \node[tool, below=0.24cm of t3] (t4) {\textbf{\textsc{Verify}} {\scriptsize+} \textbf{\textsc{Critic}}\\{\scriptsize deterministic guardrails}\\{\scriptsize + LLM judge}};

    \node[draw, dashed, rounded corners=6pt, line width=0.7pt, black!55, fit=(t1)(t2)(t3)(t4), inner sep=8pt] (toolbox) {};
    \node[lbl, anchor=south] at (toolbox.north) {Tool suite};

    \draw[barr] ([yshift=1.0cm]controller.east)  -- (t1.west);
    \draw[barr] ([yshift=0.34cm]controller.east) -- (t2.west);
    \draw[barr] ([yshift=-0.34cm]controller.east) -- (t3.west);
    \draw[barr] ([yshift=-1.0cm]controller.east) -- (t4.west);
    \node[lbl, align=center] at ($([yshift=-1.34cm]controller.east)!0.5!(t2.west)+(0,0.26)$) {call / observe};

    \node[dec, below=1.9cm of controller] (done) {Done?\\{\scriptsize round $R$ /\\ no actionable flag}};
    \node[outb, left=2.0cm of done] (output) {Corrected transcript\\ {\scriptsize + decision trace}};
    \draw[arr] ([xshift=-0.4cm]controller.south) -- ([xshift=-0.4cm]done.north);
    \draw[arr] ([xshift=0.5cm]done.north) -- node[lbl, right, align=left]{no:\\ feedback} ([xshift=0.5cm]controller.south);
    \draw[arr] (done.west) -- node[lbl, above]{yes} (output.east);

  \end{tikzpicture}
  }
  \vspace{-2mm}
  \caption{RECOVER. A \emph{controller} runs over a working state using four tools: \textsc{Retrieve} (top-$K$ entity candidates), \textsc{Fuse} (1-Best / Entity-Aware / ROVER / LLM-Select, or Adaptive Routing), \textsc{Propose} (constrained edits), and \textsc{Verify + Critic} (deterministic guardrail and LLM judge). When the critic flags an actionable missed entity, a bounded \emph{feedback} round re-retrieves for that span and re-proposes. Clean segments propose nothing and stop.}
  \label{fig:agentic}
  \vspace{-4mm}
\end{figure*}

\vspace{-1mm}

\subsection{Tool Suite}
\label{sec:tools}
\vspace{-1mm}

\paragraph{\textsc{Retrieve}.} 

Entity lists can be large (up to 6{,}198 phrases as in Table~\ref{tab:datasets}). Sending all of them to the LLM is wasteful and degrades quality, so we retrieve the top-$K$ candidates per segment with a three-signal scoring function over each phrase $b$:
\vspace{-2mm}
\begin{equation}
  \text{score}(b) = \underbrace{n_{\text{exact}} \cdot w_e}_{\text{exact token hits}}
    + \underbrace{f_{\text{best}} \cdot w_f}_{\text{fuzzy similarity}}
    + \underbrace{p_{\text{hit}} \cdot w_p}_{\text{phonetic prefix}}
\label{eq:scoring}
\vspace{-1mm}
\end{equation}
where $n_{\text{exact}}$ counts words of $b$ matching hypothesis tokens exactly, $f_{\text{best}}\!\in\![0,1]$ is the best normalized Levenshtein similarity to any hypothesis token, and $p_{\text{hit}}\!\in\!\{0,1\}$ marks a shared phonetic-key prefix (thresholds in Appendix~\ref{app:hyper}). We set $w_e\!=\!1.0$, $w_f\!=\!1.2$, $w_p\!=\!0.6$. Exact matches are the most reliable signal, fuzzy similarity is weighted highest per unit as the main mechanism for recovering corrupted entities (e.g., \textit{sitiva}$\rightarrow$\textit{cytiva}), and the phonetic cue is supplementary. The scorer returns the top $K\!=\!200$, with small lists sent in full (Section~\ref{sec:retrieval-ablation}). On a feedback round it is re-issued for the flagged span and may widen to $K_{\text{cap}}\!=\!400$ (Section~\ref{sec:orchestration}).

\vspace{-2mm}

\paragraph{\textsc{Fuse}.} It counts how many of the top-$K$ candidates appear as exact substrings in each hypothesis. These per-variant entity-hit counts drive variant selection in Entity-Aware Select and pivot selection in ROVER Ensemble.
\begin{itemize}[leftmargin=*, itemsep=1pt, topsep=2pt]
  \item \textbf{1-Best ($N\!=\!1$)}: It passes the single greedy hypothesis ($T\!=\!0$) through unchanged.
  \vspace{-2mm}
  \item \textbf{Entity-Aware Select}: It chooses the hypothesis with the maximum
  number of exact entity-candidate substring matches, with transcript length as
  a weak tie-breaker (since longer transcripts correlate with fewer deletions).
  \vspace{-2mm}
  \item \textbf{ROVER Ensemble}: It chooses an entity-aware pivot (same heuristic as Entity-Aware Select), aligns each remaining hypothesis to it with Needleman--Wunsch global alignment~\cite{needleman1970general}, and takes a majority-vote token merge (ties favour the pivot and insertions are accepted only if supported by ${\geq}3$ of~5 hypotheses).
  \vspace{-2mm}
  \item \textbf{LLM-Select}: It delegates selection to the LLM inside \textsc{Propose} by supplying all $N$ hypotheses and the top-$K$ candidates. The LLM chooses the best base variant and can propose entity-only corrections in the same call.
  \vspace{-2mm}
\item \textbf{Adaptive Routing}: It is an additional strategy that lets an LLM router pick one of the four main fusion strategies per segment from inter-hypothesis agreement and entity-hit counts.
\end{itemize}

\vspace{-1mm}

\paragraph{\textsc{Propose}.} It proposes entity-only find or replace edits via a constrained LLM which enforces strict rules that replacements must be exactly one entity-list phrase, no generic rewrites (grammar, punctuation, casing), and near-miss correction is encouraged (e.g., $\textit{citeva}$ $\rightarrow$ $\textit{cytiva}$).

\vspace{-2mm}
\paragraph{\textsc{Verify\,+\,Critic}.} Each proposed replacement passes deterministic checks before it is applied: it must be an entity-list phrase, case-only changes are discarded, and the normalized Levenshtein similarity between original and replacement must clear a floor of $0.5$, blocking unrelated substitutions (\textit{star}$\rightarrow$\textit{cytiva}) while admitting near-misses (\textit{citeva}$\rightarrow$\textit{cytiva}). Surviving edits are applied left to right, skipping overlaps. The floor is deliberately loose because the critic supplies precision, and these guardrails act as a closed-vocabulary safety gate that is never bypassed (Appendix~\ref{app:hyper}). An LLM critic then reviews the result with all $N$ hypotheses available, so it can judge from cross-hypothesis evidence. For each surviving edit it returns \emph{accept} or \emph{reject} with a reason, and it can flag a missed entity or request more candidates. Only critic-approved edits are applied, and these rejections and flags are the feedback the controller acts on.

\vspace{-1mm}
\subsection{Agentic Controller}
\vspace{-1mm}
\label{sec:orchestration}
The controller composes \textsc{Retrieve}, \textsc{Fuse}, \textsc{Propose}, and \textsc{Verify+Critic} over a working \emph{state} (transcript, candidates, accepted and rejected edits, round counter). When the critic flags a missed span as actionable, a targeted feedback round fires, bounded to $R\!=\!1$ in our experiments. It re-retrieves for that span and re-proposes with the rejected edits to avoid, the spans to focus on, and the new candidates, without recomputing fusion. This keeps feedback tied to real evidence of a recoverable entity. Section~\ref{sec:agentic-results} reports how often it fires and \emph{adds} a correction.

\vspace{-1mm}
\section{Experimental Setup}
\label{sec:experiments}
\vspace{-1mm}
\subsection{Datasets}
\vspace{-1mm}

We evaluate on five datasets spanning diverse domains (Table~\ref{tab:datasets}). \textbf{Earnings-21}~\cite{del2021earnings} comprises 44 calls segmented into $\sim$1-minute clips (2{,}086 segments). From \textbf{ATCO2}~\cite{zuluagagomez2022atco2,zuluagagomez2020asr}, we use the 1-hour test set with 446 callsign entities. \textbf{Eka-Medical}~\cite{ekacare2024medical} has 3{,}619 utterances with 6{,}198 medical entities; the \textbf{Common Voice}~\cite{ardila2020common} Corpus-22 English test set has 16{,}401 utterances, for which we extract 3{,}098 entities with a BERT NER model~\cite{thorbecke2025unifying} as it ships no entity list. \textbf{ContextASR-Bench}~\cite{wang2025contextasrbench} (English Dialogue, 5{,}273 samples) provides 3{,}704 movie-name entities.

\vspace{-1mm}
\subsection{ASR systems and LLMs}
\vspace{-1mm}
All audio is decoded with \texttt{faster-whisper}~\cite{fasterwhisper} at five temperatures $T\!\in\!\{0,0.2,0.4,0.6,0.8\}$. Our main experiments use Whisper-large-v3~\cite{radford2023robust} with GPT-4o~\cite{hurst2024gpt}, plus a GPT-4o-mini ablation (LLM-Select). To show how gains scale with recognizer size, Section~\ref{sec:main-results} reports the same methods on the weaker Whisper-small~\cite{radford2023robust}. To test whether the framework depends on Whisper-specific error structure, Section~\ref{sec:crossasr} rebuilds the Earnings-21 hypothesis set from five \emph{independent} recognizers, Whisper-large-v3 together with four commercial ASR providers (Speechmatics~\cite{speechmatics}, Deepgram~\cite{deepgram}, AssemblyAI~\cite{assemblyai}, and AWS Transcribe~\cite{awstranscribe}), each with its default English model and contributing one hypothesis. Appendix~\ref{app:openweight} additionally runs the corrector and critic on the open-weight Llama-3.3-70B~\cite{grattafiori2024llama3} and Mixtral-8x22B~\cite{jiang2024mixtral} through OpenRouter~\cite{openrouter}.

\vspace{-1mm}
\subsection{External baselines}
\vspace{-1mm}
\label{sec:baselines}
We compare against six single-pass LLM correction systems on Earnings-21, namely zero-shot and few-shot correction with task-activating prompting~\cite{yang2023generative}, N-best generative correction~\cite{ma2023n,chen2023hyporadise}, N-best selection, N-best with entity list, and direct prompting with the entity list. All are single-pass and unconstrained, with no guardrail and no critic, and all run on the same segments with the same LLM (GPT-4o-mini) and, where a vocabulary is used, the same retriever and top-$K$ candidates as RECOVER. Appendix~\ref{app:baselines} specifies each.

\begin{table}[tb]
  \centering
  \caption{Dataset statistics. Segments is the number of independently
  processed transcript units. Audio hours is the approximate total duration.
  Entities is the size of the entity phrase list. Entity Reference Tokens is the
  total number of reference words belonging to entity phrases.}
  \label{tab:datasets}
  \vspace{-2mm}
  \small
  \resizebox{\columnwidth}{!}{%
  \begin{tabular}{@{}lcccc l@{}}
    \toprule
    \textbf{Dataset} & \textbf{Segments} & \textbf{\makecell{Audio\\hours (approx.)}} & \textbf{Entities} &
    \textbf{\makecell{Entity Reference\\Tokens}} & \textbf{Entity types} \\
    \midrule
    Earnings-21
      & 2{,}086 & 38.8 & 1{,}013 & 6{,}535 & Persons, Orgs, Products \\
    ATCO2
      & 560 & 1.1 & 446 & 4{,}771 & Callsigns \\
    Eka-Medical
      & 3{,}619 & 8.4 & 6{,}198 & 42{,}449 & Medical terms \\
    Common Voice
      & 16{,}401 & 27.1 & 3{,}098 & 7{,}889 & Derived Named Entities \\
    ContextASR-Bench
      & 5{,}273 & 221.9 & 3{,}704 & 92{,}381 & Movie names \\
    \bottomrule
  \end{tabular}}
  \vspace{-4mm}
\end{table}

\vspace{-1mm}
\subsection{Evaluation}
\vspace{-1mm}
\wer{} is computed with SCLITE~\cite{fiscus1997post}. \ewer{} is computed only over reference tokens belonging to entity phrases, and relative \ewer{} reduction (RWERR) is $(\ewer_{\text{base}}-\ewer_{\text{sys}})/\ewer_{\text{base}}$. We report entity precision (P), recall (R), and F1. We also report per segment LLM calls, tokens (prompt plus completion), cost, and median latency (Section~\ref{sec:cost}).

\section{Results and Discussion}
\label{sec:results}

\begin{table*}[t]
  \centering
  \caption{Results across five datasets (all values in \%). Upper block: \textbf{Whisper-large-v3} with GPT-4o under each fusion strategy against \textbf{Baseline (no correction)}, plus a GPT-4o-mini LLM-Select ablation. Lower block: the weaker {Whisper-small} recognizer. Below each \ewer: RWERR vs.\ the corresponding baseline; below each F1: (P\,/\,R). Best \wer\,($\downarrow$), \ewer\,($\downarrow$) and F1\,($\uparrow$) per dataset in bold.}
  \label{tab:main}
  \vspace{-2mm}
  \footnotesize
  \setlength{\tabcolsep}{2.2pt}
  \renewcommand{\arraystretch}{1.1}
  \resizebox{\textwidth}{!}{%
  \begin{tabular}{@{}l l l ccc ccc ccc ccc ccc@{}}
    \toprule
    & & & \multicolumn{3}{c}{\textbf{Earnings-21}}
    & \multicolumn{3}{c}{\textbf{ATCO2}}
    & \multicolumn{3}{c}{\textbf{Eka-Medical}}
    & \multicolumn{3}{c}{\textbf{Common Voice}}
    & \multicolumn{3}{c}{\textbf{ContextASR-Bench}} \\
    \cmidrule(lr){4-6} \cmidrule(lr){7-9} \cmidrule(lr){10-12}
    \cmidrule(lr){13-15} \cmidrule(lr){16-18}
    \textbf{ASR} & \textbf{LLM} & \textbf{System}
      & \wer & \ewer & F1 & \wer & \ewer & F1 & \wer & \ewer & F1
      & \wer & \ewer & F1 & \wer & \ewer & F1 \\
    \midrule
    \multirow{7}{*}{\makecell[l]{Whisper-\\large-v3}}
    & -- & Baseline
      & 12.34 & \svs{15.88}{--} & \sv{85.29}{92.73}{78.95}
      & 35.80 & \svs{41.95}{--} & \sv{61.43}{74.87}{52.07}
      & 12.96 & \svs{15.59}{--} & \sv{90.21}{97.40}{84.02}
      & 9.96 & \svs{17.94}{--} & \sv{81.60}{97.63}{70.08}
      & 7.17 & \svs{3.19}{--} & \sv{96.30}{96.22}{96.38} \\
    \cmidrule(l){2-18}
    & {GPT-4o} & 1-Best ($N{=}1$)
      & 12.52 & \svs{9.64}{39.31} & \sv{87.97}{88.26}{87.69}
      & 39.05 & \svs{37.35}{10.97} & \sv{66.74}{72.23}{62.02}
      & 10.73 & \svs{12.07}{22.56} & \sv{92.21}{95.79}{88.88}
      & 10.25 & \svs{12.37}{31.02} & \sv{85.85}{90.59}{81.58}
      & {7.08} & \svs{2.82}{11.38} & \sv{96.35}{95.91}{96.79} \\
    \addlinespace[2pt]
    & & Entity-Aware Select
      & 12.37 & \svs{\textbf{9.50}}{\textbf{40.17}} & \sv{87.61}{87.43}{87.79}
      & 40.04 & \svs{34.96}{16.66} & \sv{68.37}{71.34}{65.63}
      & 11.46 & \svs{{12.07}}{{22.58}} & \sv{91.79}{95.21}{88.61}
      & 11.67 & \svs{10.90}{39.23} & \sv{85.99}{89.86}{82.45}
      & 7.75 & \svs{2.96}{7.03} & \sv{95.91}{95.22}{96.61} \\
    \addlinespace[2pt]
    & & ROVER Ensemble
      & \textbf{12.27} & \svs{9.59}{39.60} & \sv{87.92}{88.15}{87.69}
      & 40.54 & \svs{36.93}{11.96} & \sv{67.02}{71.51}{63.06}
      & 11.64 & \svs{12.64}{18.92} & \sv{91.71}{95.35}{88.33}
      & 11.28 & \svs{11.17}{37.74} & \sv{86.76}{91.55}{82.45}
      & 7.43 & \svs{\textbf{2.76}}{\textbf{13.25}} & \sv{96.02}{95.22}{96.82} \\
    \addlinespace[2pt]
    & & LLM-Select
      & 12.39 & \svs{9.76}{38.54} & \sv{\textbf{88.57}}{89.82}{87.36}
      & \textbf{33.70} & \svs{34.63}{17.46} & \sv{\textbf{69.76}}{79.72}{62.02}
      & \textbf{9.79} & \svs{\textbf{11.94}}{\textbf{23.41}} & \sv{\textbf{92.94}}{97.15}{89.07}
      & \textbf{9.79} & \svs{\textbf{9.16}}{\textbf{48.90}} & \sv{\textbf{87.80}}{91.71}{84.21}
      & \textbf{7.07} & \svs{2.78}{12.77} & \sv{\textbf{96.44}}{95.99}{96.88} \\
    \addlinespace[2pt]
    & & Adaptive Routing
      & 12.37 & \svs{10.02}{36.90} & \sv{87.66}{88.15}{87.17}
      & 34.98 & \svs{\textbf{34.35}}{\textbf{18.11}} & \sv{68.97}{77.29}{62.27}
      & 11.74 & \svs{12.65}{18.86} & \sv{91.36}{94.97}{88.01}
      & 11.62 & \svs{10.48}{41.55} & \sv{86.52}{89.96}{83.34}
      & 7.65 & \svs{2.96}{7.17} & \sv{95.96}{95.34}{96.60} \\
    \cmidrule(l){2-18}
    & GPT-4o-mini & LLM-Select
      & 12.48 & \svs{11.40}{28.23} & \sv{87.72}{90.25}{85.33}
      & 37.53 & \svs{41.29}{1.57} & \sv{62.76}{75.38}{53.76}
      & 11.52 & \svs{14.13}{9.38} & \sv{91.42}{97.41}{86.12}
      & 9.98 & \svs{15.57}{13.22} & \sv{83.19}{95.59}{73.63}
      & 7.27 & \svs{2.96}{7.17} & \sv{96.21}{95.76}{96.66} \\
    \midrule
    \multirow{3}{*}{\makecell[l]{Whisper-\\small}}
    & -- & Baseline
      & \textbf{13.59} & \svs{23.81}{--} & \sv{80.02}{93.25}{70.07}
      & 48.63 & \svs{51.50}{--} & \sv{54.76}{70.77}{44.66}
      & 17.23 & \svs{19.63}{--} & \sv{86.56}{96.76}{78.31}
      & 15.02 & \svs{25.57}{--} & \sv{73.40}{96.39}{59.26}
      & 9.40 & \svs{4.82}{--} & \sv{94.89}{95.26}{94.51} \\
    \cmidrule(l){2-18}
    & {GPT-4o} & LLM-Select
      & 13.63 & \svs{\textbf{14.69}}{\textbf{38.30}} & \sv{\textbf{86.12}}{90.16}{82.43}
      & \textbf{43.84} & \svs{\textbf{42.72}}{\textbf{17.05}} & \sv{\textbf{63.29}}{75.83}{54.30}
      & \textbf{13.60} & \svs{\textbf{14.55}}{\textbf{25.88}} & \sv{\textbf{90.65}}{96.65}{85.35}
      & \textbf{14.35} & \svs{\textbf{13.49}}{\textbf{47.23}} & \sv{\textbf{84.54}}{88.11}{81.24}
      & \textbf{9.26} & \svs{\textbf{3.86}}{\textbf{19.91}} & \sv{\textbf{95.24}}{95.02}{95.46} \\
    \addlinespace[2pt]
    & GPT-4o-mini & LLM-Select
      & 14.13 & \svs{17.83}{25.13} & \sv{83.72}{89.96}{78.29}
      & 52.31 & \svs{49.86}{3.17} & \sv{55.55}{69.44}{46.29}
      & 17.02 & \svs{17.68}{9.91} & \sv{88.26}{96.44}{81.37}
      & 15.83 & \svs{21.08}{17.57} & \sv{77.98}{92.19}{67.56}
      & 9.73 & \svs{4.44}{7.99} & \sv{94.70}{94.51}{94.90} \\
    \bottomrule
  \end{tabular}}
  \vspace{-4mm}
\end{table*}

\vspace{-1mm}
\subsection{Main results}
\vspace{-1mm}
\label{sec:main-results}
Table~\ref{tab:main} reports \wer, \ewer{} (RWERR beneath), and F1 (P\,/\,R beneath) over a strong Whisper-large-v3 baseline. Every GPT-4o correction method reduces \ewer{}, by up to 48.9\% relative on Common Voice. \textbf{LLM-Select} is the most consistent and strongest method: it attains the best F1 on all five datasets, the best overall \wer{} on three (ATCO2 35.80$\rightarrow$33.70, Eka-Medical 12.96$\rightarrow$9.79, Common Voice 9.96$\rightarrow$9.79), and the best or near-best \ewer{} throughout (RWERR 48.9\% on Common Voice, 38.5\% on Earnings-21, 23.4\% on Eka-Medical, 17.5\% on ATCO2 and 12.8\% on ContextASR-Bench). A practical concern is whether entity correction degrades overall transcript quality, so the choice of fusion strategy matters for preserving \wer{}. Merge-based fusion (ROVER, Entity-Aware Select) reduces \ewer{} but inflates \wer{} on the extremely noisy ATCO2 domain (35.80$\rightarrow$40.0--40.5) by injecting insertions, whereas selection-based LLM-Select keeps \wer{} below baseline. 1-Best improves over baseline on every domain but is weakest on ATCO2 (11.0\% against 17--18\%), since a single hypothesis recovers fewer of the entity deletions common in noisy ATC audio, exactly where multi-hypothesis fusion helps most. Adaptive Routing does not exceed a fixed LLM-Select, giving the best \ewer{} only on ATCO2 (Section~\ref{sec:agentic-results}).

\vspace{-1mm}

The dominant gain is entity recall (LLM-Select against baseline): $+$8.4\,pp (Earnings-21), $+$10.0\,pp (ATCO2), $+$5.1\,pp (Eka-Medical), $+$14.1\,pp (Common Voice), and $+$0.5\,pp (ContextASR-Bench, whose baseline recall is already 96.4\%), confirming that entity deletion and substitution is the main baseline weakness. Precision is largely preserved, yielding net F1 gains everywhere. GPT-4o-mini follows the same trend with smaller margins (RWERR 1.6--28.2\%), so stronger LLM reasoning translates into larger entity recovery under the same constraints. For context, \citet{huang2023contextualized} report that a contextualized attention-based encoder--decoder (AED) with decode-time biasing on Earnings-21 raises phrase-level recall from 58.66 to 79.11 ($+$20.45\,pp) and F1 from 71.50 to 79.88 (11.7\% relative) with little change in overall \wer{} (14.96$\rightarrow$15.12), but at the cost of precision (91.52 to 80.67). This is not directly comparable, since it uses a different recognizer in a decode-time contextualization setting, but our black-box post-ASR pipeline reaches comparable phrase recall on Earnings-21 (78.95$\rightarrow$87.36, $+$8.41\,pp) while keeping overall \wer{} essentially unchanged (12.34$\rightarrow$12.39) and losing less precision (92.73 to 89.82).

The lower block of Table~\ref{tab:main} repeats the Baseline and LLM-Select on the weaker Whisper-small recognizer, whose baseline is worse on every domain (\ewer{} 23.81 against 15.88 on Earnings-21) and so gives the corrector more headroom. GPT-4o LLM-Select reduces \ewer{} by 17.0--47.2\% RWERR while keeping \wer{} at or below baseline, giving larger \emph{absolute} entity-error reductions. GPT-4o-mini follows with smaller margins (3.2--25.1\%). 

Entity recall gains do not move overall \wer. Entity tokens are a small share of any transcript, 1.75\% on Earnings-21 (6{,}535 of 373{,}550 reference tokens), so even a large entity gain moves the global figure very little. Overall \wer{} should be read here as a \emph{safety} metric rather than the quantity the method optimises.


\begin{table}[tb]
  \centering
  \caption{Entity-level SCLITE alignment counts for baseline vs.\ LLM-Select (Whisper-large-v3, GPT-4o): Correct~(C), Substitution~(S), Deletion~(D), Insertion~(I), with the change $\Delta$ (LLM-Select\,$-$\,Baseline).}
  \label{tab:sclite}
  \vspace{-2mm}
  \scriptsize
  \setlength{\tabcolsep}{3pt}
  \renewcommand{\arraystretch}{0.9}
  \begin{tabular}{@{}llcccc@{}}
    \toprule
    \textbf{Dataset} & \textbf{System} & \textbf{C} & \textbf{S} & \textbf{D} & \textbf{I} \\
    \midrule
    \multirow{3}{*}{Earnings-21} & Baseline & 5{,}512 & 845 & 178 & 15 \\
      & LLM-Select & 5{,}908 & 474 & 153 & 11 \\
      & $\Delta$ & \textbf{+396} & \textbf{--371} & \textbf{--25} & \textbf{--4} \\
    \midrule
    \multirow{3}{*}{ATCO2} & Baseline & 2{,}817 & 1{,}182 & 728 & 73 \\
      & LLM-Select & 3{,}176 & 932 & 663 & 57 \\
      & $\Delta$ & \textbf{+359} & \textbf{--250} & \textbf{--65} & \textbf{--16} \\
    \midrule
    \multirow{3}{*}{Eka-Medical} & Baseline & 36{,}844 & 4{,}702 & 903 & 1{,}012 \\
      & LLM-Select & 38{,}327 & 3{,}383 & 739 & 946 \\
      & $\Delta$ & \textbf{+1{,}483} & \textbf{--1{,}319} & \textbf{--164} & \textbf{--66} \\
    \midrule
    \multirow{3}{*}{Common Voice} & Baseline & 6{,}503 & 1{,}265 & 121 & 29 \\
      & LLM-Select & 7{,}184 & 637 & 68 & 18 \\
      & $\Delta$ & \textbf{+681} & \textbf{--628} & \textbf{--53} & \textbf{--11} \\
    \midrule
    \multirow{3}{*}{ContextASR} & Baseline & 89{,}537 & 2{,}290 & 554 & 100 \\
      & LLM-Select & 89{,}875 & 2{,}053 & 453 & 62 \\
      & $\Delta$ & \textbf{+338} & \textbf{--237} & \textbf{--101} & \textbf{--38} \\
    \bottomrule
  \end{tabular}
  \vspace{-4mm}
\end{table}

Table~\ref{tab:sclite} shows that substitution reduction is the dominant correction mechanism, consistent with the constrained LLM resolving entity near-misses such as \textit{saitiva}$\rightarrow$\textit{cytiva}, \textit{left hansa}$\rightarrow$\textit{lufthansa}, and \textit{chronic holes}$\rightarrow$\textit{chronicles}. Deletions also decrease on every dataset, and insertions stay low, indicating that the system rarely adds spurious tokens.

\vspace{-1mm}
\subsection{Comparison with external baselines}
\vspace{-1mm}
\label{sec:external}
Table~\ref{tab:external} compares RECOVER with the six single-pass baselines of Section~\ref{sec:baselines} (using Whisper-large-v3, GPT-4o-mini). The entity vocabulary, not the LLM, carries most of the gain. The vocabulary-free correctors reach 11.3--13.0\% relative \ewer{} reduction against 28--33\% for every system given the retrieved candidates, and N-best selection, which may re-rank but not edit, is essentially flat (0.4\%). Direct prompting with the same entity list reaches 10.60 against 11.40 with RECOVER (LLM-Select, GPT-4o-mini). RECOVER wins instead on the scope of the edit, which \ewer{} does not expose. Entity tokens are only 1.75\% of this corpus, yet direct prompting changes 5.85\% of tokens and N-best-with-entities 7.93\%, against RECOVER's 0.21\%, and relative to the uncorrected baseline they add 1{,}502 and 2{,}530 overall deletions against RECOVER's 396. The unconstrained systems buy part of their entity accuracy by dropping other content, which overall \wer{} hides.

\begin{table}[tb]
  \centering
  \caption{External baselines on Earnings-21 (Whisper-large-v3, GPT-4o-mini). $\Delta$tok is the fraction of tokens changed against the input hypothesis (net word change in parentheses).}
  \label{tab:external}
  \vspace{-2mm}
  \scriptsize
  \setlength{\tabcolsep}{2pt}
  \renewcommand{\arraystretch}{1.0}
  \begin{tabular}{@{}lccc@{}}
    \toprule
    \textbf{System} & \textbf{\wer} & \textbf{\ewer\ (RWERR)} & \textbf{$\Delta$tok (net)} \\
    \midrule
    Baseline (no correction) & 12.34 & 15.88 (--) & -- \\
    \midrule
    Zero-shot correction & 12.84 & 14.09 (11.3) & -- \\
    Few-shot correction & 12.65 & 14.05 (11.6) & -- \\
    N-best generative corr. & 13.10 & 13.82 (13.0) & -- \\
    N-best selection & 12.54 & 15.82 (0.4) & -- \\
    \midrule
    N-best w/ entity list & 12.82 & 10.97 (30.9) & 7.93\% ($-$3{,}271) \\
    Direct prompt w/ entity list & 12.55 & \textbf{10.60 (33.2)} & 5.85\% ($-$1{,}213) \\
    RECOVER (LLM-Select) & \textbf{12.48} & 11.40 (28.2) & \textbf{0.21\% ($-$28)} \\
    \bottomrule
  \end{tabular}
  \vspace{-4mm}
\end{table}

\vspace{-1mm}
\subsection{Hypotheses from independent recognizers}
\vspace{-1mm}
\label{sec:crossasr}
Every result so far draws its $N$ hypotheses from temperature sampling of one recognizer. On Earnings-21, RECOVER consumes one hypothesis from each of five independent recognizers (Table~\ref{tab:crossasr}). With GPT-4o, LLM-Select reaches 7.96 \ewer{} and ROVER Ensemble 7.44, a 53.2\% relative reduction over the Whisper-large-v3 baseline and 45.8\% better on entities than the strongest provider (AssemblyAI, 13.73), while overall \wer{} (9.24) does not beat the best single provider (AWS Transcribe, 9.11), as expected for a system that edits only entity spans. ROVER Ensemble, which inflated \wer{} under temperature sampling (Section~\ref{sec:main-results}), is strongest here, since a token-level merge pays off when hypotheses are independent.

\begin{table}[tb]
  \centering
  \caption{Earnings-21 with one hypothesis from each of five independent recognizers. RECOVER is unmodified. RWERR over the Whisper-large-v3 baseline (no correction) in parentheses.}
  \label{tab:crossasr}
  \vspace{-2mm}
  \scriptsize
  \setlength{\tabcolsep}{6pt}
  \begin{tabular}{@{}lcc@{}}
    \toprule
    \textbf{System} & \textbf{\wer} & \textbf{\ewer} \\
    \midrule
    Whisper-large-v3 & 12.34 & 15.88 \\
    Speechmatics & 10.07 & 16.47 \\
    Deepgram & 14.70 & 13.93 \\
    AssemblyAI & 11.68 & \textbf{13.73} \\
    AWS Transcribe & \textbf{9.11} & 22.97 \\
    \midrule
    5 rec.\ $+$ RECOVER (GPT-4o-mini) & 11.54 & 9.38 (40.9) \\
    5 rec.\ $+$ LLM-Select (GPT-4o) & 11.56 & 7.96 (49.9) \\
    5 rec.\ $+$ ROVER Ens.\ (GPT-4o) & 9.24 & \textbf{7.44 (53.2)} \\
    \bottomrule
  \end{tabular}
  \vspace{-2mm}
\end{table}

\begin{table}[tb]
  \centering
  \caption{Entity-list coverage and noise on Earnings-21 (Whisper-large-v3, GPT-4o-mini). Coverage is the fraction of the 1{,}013-phrase list retained. The last column adds 769 distractors.}
  \label{tab:listcov}
  \vspace{-2mm}
  \scriptsize
  \setlength{\tabcolsep}{4pt}
  \begin{tabular}{@{}lccccc@{}}
    \toprule
    \textbf{List coverage} & \textbf{100\%} & \textbf{75\%} & \textbf{50\%} & \textbf{25\%} & \textbf{$+$76\% distr.} \\
    \midrule
    \ewer\,$\downarrow$ & \textbf{11.40} & 12.23 & 13.34 & 14.60 & 11.78 \\
    RWERR\,$\uparrow$ & \textbf{28.2} & 23.0 & 16.0 & 8.1 & 25.8 \\
    \wer & 12.48 & 12.49 & 12.48 & 12.51 & 12.48 \\
    \bottomrule
  \end{tabular}
  \vspace{-4mm}
\end{table}

\vspace{-1mm}
\subsection{Entity-list coverage and noise}
\vspace{-1mm}
\label{sec:listcoverage}
RECOVER requires a domain entity list, so its behaviour when that list is incomplete or noisy decides whether it is deployable. We degrade the 1{,}013-phrase Earnings-21 list, and separately enlarge it with the 769 distractors of \citet{fox2022improving} (construction in Appendix~\ref{app:listcov}). No setting pushes the entity numbers below the uncorrected baseline, and overall \wer{} stays inside 12.48--12.51 throughout (Table~\ref{tab:listcov}). This follows from the closed-vocabulary constraint: because every applied edit must be a list phrase, an entity absent from the list is a missed opportunity rather than a new error, which is what makes an incomplete list safe to deploy. Similarly, noise is cheap, since the retriever ranks distractors below phrases that match the hypotheses.

\vspace{-1mm}
\subsection{Retrieval: candidate budget and scoring weights}
\vspace{-1mm}
\label{sec:retrieval-ablation}

\emph{Candidate budget $K$.} \ewer{} improves from $K\!=\!50$ to a minimum at $K\!=\!200$ (11.46, 27.8\% RWERR) and slightly worsens at $K\!=\!400$ and $K\!=\!\text{max}$ (full 1{,}013-entity list, 11.71). Adding candidates beyond $\sim$200 does not help, while $K\!=\!200$ is already large enough to cover the entities relevant to the hypotheses. In Table~\ref{tab:retrieval}, $K$ is held fixed even on feedback re-retrieval to isolate its effect, whereas in the main experiments $K$ is raised to $K_{\text{cap}}\!=\!400$ on feedback. Raising $K$ only during feedback gives a small additional gain over a constant $K\!=\!200$ (Appendix~\ref{app:kfeedback}).

\emph{Scoring weights.} With $K\!=\!200$, the chosen weighting $(w_e,w_f,w_p)\!=\!(1,1.2,0.6)$ is best (11.46). Any single signal is markedly worse (exact-only 13.10, phonetic-only 13.93), and both uniform weights (1,1,1: 11.60) and fuzzy-only (0,1,0: 11.71) trail it. This confirms that combining exact, fuzzy, and phonetic signals helps, and that up-weighting fuzzy similarity is the right choice. A finer grid around the chosen setting is in Appendix~\ref{app:weightgrid}.

\begin{table}[tb]
  \centering
  \caption{Retrieval on Earnings-21 (Whisper-large-v3, GPT-4o-mini, LLM-Select): candidate budget $K$ and scoring weights (at $K\!=\!200$). $K$ is held fixed on feedback re-retrieval here.}
  \label{tab:retrieval}
  \vspace{-2mm}
  \scriptsize
  \setlength{\tabcolsep}{3pt}
  \begin{tabular}{@{}llc@{}}
    \toprule
    \textbf{Factor} & \textbf{Setting} & \textbf{\ewer\,$\downarrow$ (RWERR)} \\
    \midrule
    \multirow{5}{*}{Budget $K$}
      & 50 & 11.60 (27.0) \\
      & 100 & 11.52 (27.5) \\
      & \textbf{200} & \textbf{11.46 (27.8)} \\
      & 400 & 11.64 (26.7) \\
      & max ($|\mathcal{E}|{=}1013$) & 11.71 (26.3) \\
    \midrule
    \multirow{5}{*}{\makecell[l]{Weights\\$(w_e,w_f,w_p)$}}
      & \textbf{$(1,1.2,0.6)$} & \textbf{11.46 (27.8)} \\
      & $(1,1,1)$ uniform & 11.60 (27.0) \\
      & $(0,1,0)$ fuzzy-only & 11.71 (26.3) \\
      & $(1,0,0)$ exact-only & 13.10 (17.5) \\
      & $(0,0,1)$ phonetic-only & 13.93 (12.3) \\
    \bottomrule
  \end{tabular}
  \vspace{-2mm}
\end{table}

\subsection{Feedback, guardrails, and the critic}
\vspace{-1mm}
\label{sec:agentic-results}

\begin{table}[tb]
  \centering
  \caption{Feedback and verification statistics for RECOVER (LLM-Select, Whisper-large-v3, GPT-4o), per dataset. Feedback\% = segments that triggered a feedback round, Helped(seg)\% / Helped(fb)\% = fraction of \emph{all} segments / of \emph{feedback} segments where feedback added a correction, G-rej / C-rej = total edits rejected by the deterministic guardrail or the LLM critic.}
  \label{tab:process}
  \vspace{-2mm}
  \scriptsize
  \setlength{\tabcolsep}{3.5pt}
  \resizebox{\columnwidth}{!}{%
  \begin{tabular}{@{}lccccc@{}}
    \toprule
    \textbf{Dataset} & \textbf{Feedback\%} & \textbf{Helped(seg)\%} & \textbf{Helped(fb)\%} & \textbf{G-rej} & \textbf{C-rej} \\
    \midrule
    Earnings-21 & 5.6 & 1.2 & 21.4 & 2{,}297 & 112  \\
    ATCO2 & 2.3 & 0.5 & 23.1 & 410 & 34  \\
    Eka-Medical & 5.2 & 0.8 & 15.4 & 2{,}873 & 202  \\
    Common Voice & 2.2 & 0.2 & 9.1 & 6{,}035 & 557  \\
    ContextASR-Bench & 12.3 & 4.6 & 37.7 & 6{,}865 & 632 \\
    \bottomrule
  \end{tabular}}
  \vspace{-4mm}
\end{table}

Every RECOVER configuration in Table~\ref{tab:main} passes through a deterministic guardrail, an LLM critic, and up to one bounded feedback round. Table~\ref{tab:process} reports how often feedback fires and how the two verification layers behave. Feedback is rare and targeted, firing on 2.2--12.3\% of segments and adding a correction on 9.1--37.7\% of those invocations. The two layers do complementary work: the guardrail rejects the bulk of invalid proposals (410--6{,}865) as a high-volume safety filter, while the critic rejects a much smaller residue that passed the gate (34--632) as a precision check. Disabling both leaves overall \wer{} essentially unchanged ($+0.01$\,pp) but costs 2.5 points of relative \ewer{} reduction (Appendix~\ref{app:criticwer}), so they buy entity precision rather than transcript quality. Under adaptive routing the router mostly picks Entity-Aware Select, which is why it does not exceed a fixed LLM-Select (Appendix~\ref{app:routing}).

\begin{table}[tb]
  \centering
  \caption{Cost and latency on Earnings-21 (Whisper-large-v3). Tokens are prompt\,$+$\,completion per segment, cost uses list API prices, latency is the median per-segment total across all LLM calls for that segment.}
  \label{tab:cost}
  \vspace{-2mm}
  \scriptsize
  \setlength{\tabcolsep}{2pt}
  \resizebox{\columnwidth}{!}{%
  \begin{tabular}{@{}lcccc@{}}
    \toprule
    \textbf{Configuration} & \textbf{calls/seg} & \textbf{tokens/seg} & \textbf{USD/1k seg} & \textbf{median s/seg} \\
    \midrule
    Single-pass zero-shot, GPT-4o-mini & 1.00 & 277$+$192 & 0.16 & 3.09 \\
    Single-pass $+$ entity list, GPT-4o-mini & 1.00 & 1222$+$186 & 0.29 & 4.37 \\
    RECOVER, GPT-4o & 2.11 & 4096$+$333 & 13.57 & 6.14 \\
    RECOVER, GPT-4o-mini & 2.35 & 4488$+$277 & 0.84 & 5.57 \\
    \bottomrule
  \end{tabular}}
  \vspace{-4mm}
\end{table}

\vspace{-1mm}
\subsection{Cost and latency}
\vspace{-1mm}
\label{sec:cost}
Table~\ref{tab:cost} reports the costs: RECOVER issues 2.11 LLM calls per segment with GPT-4o against one for a prompt-only corrector, and the multi-hypothesis prompt dominates. The five-hypothesis fusion call carries 4{,}096 prompt tokens against 277 for a single-hypothesis zero-shot pass, while feedback fires on 5.6\% of segments for $+0.11$ calls and about 5\% of cost (Appendix~\ref{app:cost}). Model choice dominates the cost: GPT-4o-mini runs the same pipeline 16 times cheaper while retaining 28.2\% relative \ewer{} reduction against 38.5\% with GPT-4o (Table~\ref{tab:main}). Latency tracks calls rather than model size. A cost-constrained deployment should therefore change the LLM and a latency-constrained one disable the critic and feedback (Appendix~\ref{app:criticwer}), rather than give up the multi-hypothesis evidence. Open-weight correctors are a further option (Appendix~\ref{app:openweight}).

\vspace{-1mm}
\section{Conclusion}
\label{sec:conclusion}
\vspace{-2mm}
We presented RECOVER, a post-ASR entity-correction framework that retrieves entity candidates, fuses multiple ASR hypotheses, proposes constrained edits, filters them through deterministic guardrails, and verifies them with an LLM critic that can trigger a bounded feedback round. Over a strong Whisper-large-v3 baseline it yields up to 49\% relative \ewer{} reduction across five domains, and up to 53\% on Earnings-21 using hypotheses from five independent recognizers. The findings that underpin these results are that alternative hypotheses of the same audio can carry entity evidence that the single hypothesis may have lost, and constraining every edit to a retrieved vocabulary is what keeps the correction targeted. Since the framework is agnostic to how those hypotheses are produced and needs no access to the recognizer, it applies to any black-box or commercial ASR.

\newpage
\section*{Limitations}
RECOVER assumes access to a domain entity list. Section~\ref{sec:listcoverage} quantifies that dependence rather than removing it, that is, at 25\% coverage the relative \ewer{} reduction falls from 28.2 to 8.1, so the method is bounded by the quality of the list it is given.

Direct prompting with the same retrieved entity list reaches 10.60 against 11.40 with RECOVER on Earnings-21 (Section~\ref{sec:external}). Our claim is the trade-off, a comparable entity gain at a 0.21\% rather than 5.85\% edit footprint. A deployment that does not value transcript preservation should prefer the simpler baseline.

The orchestration is not free. The critic adds an LLM call per segment (and a routing call under adaptive routing), and RECOVER issues 2.11 against 1.00 calls per segment, roughly three times the dollar cost of a single-pass corrector carrying the same entity list on the same LLM (Section~\ref{sec:cost}). The feedback round in particular is better understood as an optional precision setting: it fires on 2.2--12.3\% of segments and adds a correction on only 0.2--4.6\% of all segments, and disabling it together with the critic costs about 2.5 points of relative \ewer{} reduction on Earnings-21 (Appendix~\ref{app:criticwer}). The deterministic guardrail, not the feedback loop, does the bulk of the safety work.

Adaptive routing does not exceed a fixed LLM-Select on any dataset except ATCO2, where the margin is marginal, and the router overwhelmingly selects Entity-Aware Select (Appendix~\ref{app:routing}). Richer routing signals or a trained router would be needed to make it a contribution rather than an ablation.

Multi-hypothesis diversity from temperature sampling is recognizer-dependent and may shrink as ASR quality improves (Table~\ref{tab:main}), although independent recognizers supply substantially more evidence where several are available (Section~\ref{sec:crossasr}). Finally, our cost figures are hosted list prices at the time of writing and are not the cost of self-hosting.

\section*{Ethical Considerations}
RECOVER targets high-stakes domains such as finance, medicine, and air-traffic control, where both a wrong correction and a missed entity can have real consequences. The closed-vocabulary constraint and the LLM critic are designed to suppress hallucinated edits, and the deterministic guardrail never applies an out-of-list phrase, but residual errors remain. We therefore view RECOVER as a tool to assist human review rather than a system to be relied on without oversight in safety-critical settings such as clinical documentation or air-traffic control. We evaluate on publicly available or licensed datasets (Earnings-21, ATCO2, Eka-Medical, Common Voice, ContextASR-Bench). Speech in the medical and air-traffic domains can contain sensitive or personally identifiable information, and we use these datasets only for evaluation under their respective terms and report aggregate metrics only. Our pipeline also relies on a third-party LLM and on multiple hypotheses, which carry the usual model-bias risks and send transcripts to an external service.

The entity list is itself a source of bias. Every applied edit must be a phrase in the supplied vocabulary, so names absent from that list are systematically never recovered, and the same guardrail that suppresses hallucination also fixes this asymmetry in place. If a list under-represents names from particular languages, regions, or communities, the system improves recognition for the names it contains and leaves the others at baseline. Section~\ref{sec:listcoverage} shows this quantitatively, as coverage bounds the achievable gain almost proportionally. Choosing what goes into the entity list is therefore a fairness decision and a list should be checked before deployment for whether it covers the names of the people the system will serve. Our results are averages where the corpora we use do not reliably label speakers, so we cannot report errors by speaker group, and our evaluation therefore cannot show whether the system treats all groups equally.

The framework issues 2.00 -- 2.35 LLM calls and 4.4 -- 4.8k tokens per segment (Section~\ref{sec:cost}), several times the token footprint of a single-pass corrector, so it carries a correspondingly larger energy and carbon cost that a deployment should weigh against the entity gain. We report calls, tokens, cost, and latency for every configuration so that this trade-off is explicit. The cheapest configuration we measured, the open-weight Llama-3.3-70B corrector at USD 0.60 per 1{,}000 segments (Appendix~\ref{app:openweight}), also attains a larger relative reduction (30.0\%) than GPT-4o-mini, and disabling the critic and feedback removes roughly one call per segment at a small accuracy cost.


\newpage

\bibliography{recover}

@inproceedings{radford2023robust,
  title={Robust speech recognition via large-scale weak supervision},
  author={Radford, Alec and Kim, Jong Wook and Xu, Tao and Brockman, Greg and McLeavey, Christine and Sutskever, Ilya},
  booktitle={International conference on machine learning},
  pages={28492--28518},
  year={2023},
  organization={PMLR}
}

@article{sainath2019two,
  title={Two-pass end-to-end speech recognition},
  author={Sainath, Tara N and Pang, Ruoming and Rybach, David and He, Yanzhang and Prabhavalkar, Rohit and Li, Wei and Visontai, Mirk{\'o} and Liang, Qiao and Strohman, Trevor and Wu, Yonghui and others},
  journal={arXiv preprint arXiv:1908.10992},
  year={2019}
}

@article{huang2023contextualized,
  title={Contextualized end-to-end speech recognition with contextual phrase prediction network},
  author={Huang, Kaixun and Zhang, Ao and Yang, Zhanheng and Guo, Pengcheng and Mu, Bingshen and Xu, Tianyi and Xie, Lei},
  journal={arXiv preprint arXiv:2305.12493},
  year={2023}
}

@article{fox2022improving,
  title={Improving contextual recognition of rare words with an alternate spelling prediction model},
  author={Fox, Jennifer Drexler and Delworth, Natalie},
  journal={arXiv preprint arXiv:2209.01250},
  year={2022}
}

@article{del2021earnings,
  title={Earnings-21: A practical benchmark for asr in the wild},
  author={Del Rio, Miguel and Delworth, Natalie and Westerman, Ryan and Huang, Michelle and Bhandari, Nishchal and Palakapilly, Joseph and McNamara, Quinten and Dong, Joshua and Zelasko, Piotr and Jett{\'e}, Miguel},
  journal={arXiv preprint arXiv:2104.11348},
  year={2021}
}

@article{ma2025asr,
  title={Asr error correction using large language models},
  author={Ma, Rao and Qian, Mengjie and Gales, Mark and Knill, Kate},
  journal={IEEE Transactions on Audio, Speech and Language Processing},
  year={2025},
  publisher={IEEE}
}

@inproceedings{thorbecke2025unifying,
  title={Unifying Global and Near-Context Biasing in a Single Trie Pass},
  author={Thorbecke, Iuliia and Villatoro-Tello, Esa{\'u} and Zuluaga, Juan Pablo and Kumar, Shashi and Burdisso, Sergio and Rangappa, Pradeep and Carofilis, Andr{\'e}s and Madikeri, Srikanth and Motlicek, Petr and Pandia, Karthik and others},
  booktitle={International Conference on Text, Speech, and Dialogue},
  pages={170--181},
  year={2025},
  organization={Springer}
}

@article{derouiche2025agentic,
  title={Agentic ai frameworks: Architectures, protocols, and design challenges},
  author={Derouiche, Hana and Brahmi, Zaki and Mazeni, Haithem},
  journal={arXiv preprint arXiv:2508.10146},
  year={2025}
}

@inproceedings{fiscus1997post,
  title={A post-processing system to yield reduced word error rates: Recognizer output voting error reduction (ROVER)},
  author={Fiscus, Jonathan G},
  booktitle={1997 IEEE Workshop on Automatic Speech Recognition and Understanding Proceedings},
  pages={347--354},
  year={1997},
  organization={IEEE}
}

@inproceedings{yao2022react,
  title={React: Synergizing reasoning and acting in language models},
  author={Yao, Shunyu and Zhao, Jeffrey and Yu, Dian and Du, Nan and Shafran, Izhak and Narasimhan, Karthik and Cao, Yuan},
  booktitle={The Eleventh International Conference on Learning Representations (ICLR)},
  year={2023}
}

@article{baevski2020wav2vec,
  title={wav2vec 2.0: A framework for self-supervised learning of speech representations},
  author={Baevski, Alexei and Zhou, Yuhao and Mohamed, Abdelrahman and Auli, Michael},
  journal={Advances in neural information processing systems},
  volume={33},
  pages={12449--12460},
  year={2020}
}

@article{needleman1970general,
  title={A general method applicable to the search for similarities in the amino acid sequence of two proteins},
  author={Needleman, Saul B and Wunsch, Christian D},
  journal={Journal of molecular biology},
  volume={48},
  number={3},
  pages={443--453},
  year={1970},
  publisher={Elsevier}
}

@article{jain2020contextual,
  title={Contextual RNN-T for open domain ASR},
  author={Jain, Mahaveer and Keren, Gil and Mahadeokar, Jay and Zweig, Geoffrey and Metze, Florian and Saraf, Yatharth},
  journal={arXiv preprint arXiv:2006.03411},
  year={2020}
}

@article{le2021contextualized,
  title={Contextualized streaming end-to-end speech recognition with trie-based deep biasing and shallow fusion},
  author={Le, Duc and Jain, Mahaveer and Keren, Gil and Kim, Suyoun and Shi, Yangyang and Mahadeokar, Jay and Chan, Julian and Shangguan, Yuan and Fuegen, Christian and Kalinli, Ozlem and others},
  journal={arXiv preprint arXiv:2104.02194},
  year={2021}
}

@inproceedings{yang2023generative,
  title={Generative speech recognition error correction with large language models and task-activating prompting},
  author={Yang, Chao-Han Huck and Gu, Yile and Liu, Yi-Chieh and Ghosh, Shalini and Bulyko, Ivan and Stolcke, Andreas},
  booktitle={2023 IEEE Automatic Speech Recognition and Understanding Workshop (ASRU)},
  pages={1--8},
  year={2023},
  organization={IEEE}
}

@article{pu2023multi,
  title={Multi-stage large language model correction for speech recognition},
  author={Pu, Jie and Nguyen, Thai-Son and St{\"u}ker, Sebastian},
  journal={arXiv preprint arXiv:2310.11532},
  year={2023}
}

@inproceedings{manakul2023selfcheckgpt,
  title={Selfcheckgpt: Zero-resource black-box hallucination detection for generative large language models},
  author={Manakul, Potsawee and Liusie, Adian and Gales, Mark},
  booktitle={Proceedings of the 2023 conference on empirical methods in natural language processing},
  pages={9004--9017},
  year={2023}
}

@article{ma2023n,
  title={N-best t5: Robust asr error correction using multiple input hypotheses and constrained decoding space},
  author={Ma, Rao and Gales, Mark JF and Knill, Kate M and Qian, Mengjie},
  journal={arXiv preprint arXiv:2303.00456},
  year={2023}
}

@inproceedings{azmi2025llm,
  title={LLM-Enhanced Spoken Named Entity Recognition Leveraging ASR N-Best Hypotheses},
  author={Azmi, Farhan and Tong, Rong},
  booktitle={2025 International Conference on Asian Language Processing (IALP)},
  pages={153--158},
  year={2025},
  organization={IEEE}
}

@article{xu2025large,
  title={Large language models based asr error correction for child conversations},
  author={Xu, Anfeng and Feng, Tiantian and Kim, So Hyun and Bishop, Somer and Lord, Catherine and Narayanan, Shrikanth},
  journal={arXiv preprint arXiv:2505.16212},
  year={2025}
}

@article{gulati2020conformer,
  title={Conformer: Convolution-augmented transformer for speech recognition},
  author={Gulati, Anmol and Qin, James and Chiu, Chung-Cheng and Parmar, Niki and Zhang, Yu and Yu, Jiahui and Han, Wei and Wang, Shibo and Zhang, Zhengdong and Wu, Yonghui and others},
  journal={arXiv preprint arXiv:2005.08100},
  year={2020}
}

@inproceedings{burchi2021efficient,
  title={Efficient conformer: Progressive downsampling and grouped attention for automatic speech recognition},
  author={Burchi, Maxime and Vielzeuf, Valentin},
  booktitle={2021 IEEE Automatic Speech Recognition and Understanding Workshop (ASRU)},
  pages={8--15},
  year={2021},
  organization={IEEE}
}

@article{hurst2024gpt,
  title={Gpt-4o system card},
  author={Hurst, Aaron and Lerer, Adam and Goucher, Adam P and Perelman, Adam and Ramesh, Aditya and Clark, Aidan and Ostrow, AJ and Welihinda, Akila and Hayes, Alan and Radford, Alec and others},
  journal={arXiv preprint arXiv:2410.21276},
  year={2024}
}

@article{zuluagagomez2022atco2,
  title={ATCO2 corpus: A large-scale dataset for research on automatic speech recognition and natural language understanding of air traffic control communications},
  author={Zuluaga-Gomez, Juan and Vesel{\'y}, Karel and Sz{\"o}ke, Igor and Blatt, Alexander and Motlicek, Petr and Kocour, Martin and Rigault, Mickael and Choukri, Khalid and Prasad, Amrutha and Sarfjoo, Seyyed Saeed and Nigmatulina, Iuliia and Cevenini, Claudia and Kol{\v{c}}{\'a}rek, Pavel and Tart, Allan and {\v{C}}ernock{\'y}, Jan and Klakow, Dietrich},
  journal={arXiv preprint arXiv:2211.04054},
  year={2022}
}

@inproceedings{zuluagagomez2020asr,
  title={Automatic Speech Recognition Benchmark for Air-Traffic Communications},
  author={Zuluaga-Gomez, Juan and Motlicek, Petr and Zhan, Qingran and Vesely, Karel and Braun, Rudolf},
  booktitle={Proc. Interspeech},
  pages={2297--2301},
  year={2020}
}

@inproceedings{ardila2020common,
  title={Common Voice: A Massively-Multilingual Speech Corpus},
  author={Ardila, Rosana and Branson, Megan and Davis, Kelly and Henretty, Michael and Kohler, Michael and Meyer, Josh and Morais, Reuben and Saunders, Lindsay and Tyers, Francis M and Weber, Gregor},
  booktitle={Proceedings of the Twelfth Language Resources and Evaluation Conference},
  pages={4218--4222},
  year={2020}
}

@article{wang2025contextasrbench,
  title={ContextASR-Bench: A Massive Contextual Speech Recognition Benchmark},
  author={Wang, He and Ma, Linhan and Guo, Dake and Wang, Xiong and Xie, Lei and Xu, Jin and Lin, Junyang},
  journal={arXiv preprint arXiv:2507.05727},
  year={2025}
}

@misc{fasterwhisper,
  title  = {faster-whisper},
  author = {{SYSTRAN}},
  year   = {2023},
  howpublished = {\url{https://github.com/SYSTRAN/faster-whisper}}
}

@inproceedings{wang2024contextbiasing,
  title     = {Contextual Biasing with the Knuth-Morris-Pratt Matching Algorithm},
  author    = {Wang, Weiran and Wu, Zelin and Caseiro, Diamantino and Munkhdalai, Tsendsuren and Sim, Khe Chai and Rondon, Pat and Pundak, Golan and Song, Gan and Prabhavalkar, Rohit and Meng, Zhong and Zhao, Ding and Sainath, Tara and He, Yanzhang and Moreno Mengibar, Pedro},
  booktitle = {Proc. Interspeech 2024},
  pages     = {282--286},
  year      = {2024},
  doi       = {10.21437/Interspeech.2024-1349}
}

@inproceedings{panayotov2015librispeech,
  title     = {Librispeech: An {ASR} corpus based on public domain audio books},
  author    = {Panayotov, Vassil and Chen, Guoguo and Povey, Daniel and Khudanpur, Sanjeev},
  booktitle = {Proc. IEEE International Conference on Acoustics, Speech and Signal Processing (ICASSP)},
  pages     = {5206--5210},
  year      = {2015}
}

@article{ayache2024whispener,
  title   = {{WhisperNER}: Unified Open Named Entity and Speech Recognition},
  author  = {Ayache, Gil and Pirchi, Menachem and Navon, Aviv and Shamsian, Aviv and Hetz, Gill and Keshet, Joseph},
  journal = {arXiv preprint arXiv:2409.08107},
  year    = {2024}
}

@article{grattafiori2024llama3,
  title   = {The {Llama} 3 Herd of Models},
  author  = {Grattafiori, Aaron and Dubey, Abhimanyu and Jauhri, Abhinav and Pandey, Abhinav and Kadian, Abhishek and Al-Dahle, Ahmad and others},
  journal = {arXiv preprint arXiv:2407.21783},
  year    = {2024}
}

@article{jiang2024mixtral,
  title   = {Mixtral of Experts},
  author  = {Jiang, Albert Q. and Sablayrolles, Alexandre and Roux, Antoine and Mensch, Arthur and Savary, Blanche and Bamford, Chris and Chaplot, Devendra Singh and de las Casas, Diego and Hanna, Emma Bou and Bressand, Florian and others},
  journal = {arXiv preprint arXiv:2401.04088},
  year    = {2024}
}

@misc{openrouter,
  title        = {{OpenRouter}: A unified interface for {LLM}s},
  author       = {{OpenRouter}},
  year         = {2025},
  howpublished = {\url{https://openrouter.ai}}
}

@article{chen2023hyporadise,
  title   = {HyPoradise: An Open Baseline for Generative Speech Recognition with Large Language Models},
  author  = {Chen, Chen and Hu, Yuchen and Yang, Chao-Han Huck and Siniscalchi, Sabato Marco and Chen, Pin-Yu and Chng, Eng Siong},
  journal = {Advances in Neural Information Processing Systems (NeurIPS)},
  year    = {2023}
}

@inproceedings{schick2023toolformer,
  title     = {Toolformer: Language Models Can Teach Themselves to Use Tools},
  author    = {Schick, Timo and Dwivedi-Yu, Jane and Dess{\`i}, Roberto and Raileanu, Roberta and Lomeli, Maria and Hambro, Eric and Zettlemoyer, Luke and Cancedda, Nicola and Scialom, Thomas},
  booktitle = {Advances in Neural Information Processing Systems (NeurIPS)},
  year      = {2023}
}

@article{shinn2023reflexion,
  title   = {Reflexion: Language Agents with Verbal Reinforcement Learning},
  author  = {Shinn, Noah and Cassano, Federico and Berman, Edward and Gopinath, Ashwin and Narasimhan, Karthik and Yao, Shunyu},
  journal = {Advances in Neural Information Processing Systems (NeurIPS)},
  year    = {2023}
}

@article{wang2024survey,
  title   = {A Survey on Large Language Model based Autonomous Agents},
  author  = {Wang, Lei and Ma, Chen and Feng, Xueyang and Zhang, Zeyu and Yang, Hao and Zhang, Jingsen and Chen, Zhiyuan and Tang, Jiakai and Chen, Xu and Lin, Yankai and others},
  journal = {Frontiers of Computer Science},
  volume  = {18},
  number  = {6},
  year    = {2024}
}

@misc{speechmatics,
  title        = {Speechmatics Speech-to-Text {API}},
  author       = {{Speechmatics}},
  year         = {2026},
  howpublished = {\url{https://www.speechmatics.com}}
}

@misc{deepgram,
  title        = {Deepgram Speech-to-Text {API}},
  author       = {{Deepgram}},
  year         = {2026},
  howpublished = {\url{https://deepgram.com}}
}

@misc{assemblyai,
  title        = {{AssemblyAI} Speech-to-Text {API}},
  author       = {{AssemblyAI}},
  year         = {2026},
  howpublished = {\url{https://www.assemblyai.com}}
}

@misc{awstranscribe,
  title        = {Amazon Transcribe},
  author       = {{Amazon Web Services}},
  year         = {2026},
  howpublished = {\url{https://aws.amazon.com/transcribe}}
}

@misc{ekacare2024medical,
  title        = {Eka Medical {ASR} Evaluation Dataset},
  author       = {{Eka Care}},
  year         = {2024},
  howpublished = {\url{https://huggingface.co/datasets/ekacare/eka-medical-asr-evaluation-dataset}}
}

\newpage

\appendix

\section{External Baseline Configurations}
\label{app:baselines}
All six baselines in Table~\ref{tab:external} are run on Earnings-21, Whisper-large-v3 hypotheses, with GPT-4o-mini as the LLM. None of them applies the deterministic guardrail or the critic, and where a vocabulary is used it is the same retriever and the same top-$K$ candidate set that RECOVER receives.

\begin{itemize}[leftmargin=*, itemsep=1pt, topsep=2pt]
  \item \textbf{Zero-shot correction.} The 1-Best transcript is given to the LLM with an instruction to correct recognition errors and return the corrected transcript, with no examples and no vocabulary~\cite{yang2023generative}.
  \item \textbf{Few-shot correction (task-activating prompting).} As above, with an error-pattern that describes the kinds of ASR error to expect and three worked near-miss examples, following the task-activating prompting recipe of~\citet{yang2023generative}.
  \item \textbf{N-best generative correction.} All five hypotheses are supplied at once and the LLM writes a fresh transcript from their disagreement, with no vocabulary~\cite{ma2023n,chen2023hyporadise}.
  \item \textbf{N-best selection.} The N-best-constrained variant: the LLM returns the index of one of the five hypotheses and may not edit it. This isolates re-ranking from correction.
  \item \textbf{N-best with entity list.} The generative N-best prompt plus retrieved top-200 entity candidates, with free-form output and no constraint enforcement.
  \item \textbf{Direct prompting with the entity list.} The 1-Best transcript plus the same retrieved top-200 candidates, free-form output, no guardrail and no critic.
\end{itemize}

Edit scope in Table~\ref{tab:external} is measured by aligning each system's output against its own input hypothesis and counting the fraction of tokens that changed. The net word figure is the change in total transcript length. Deletion counts are read from the SCLITE alignment against the reference.

\section{Hyperparameters}
\label{app:hyper}

\paragraph{Retrieval scoring (Eq.~\ref{eq:scoring}).}
The score combines three standard signals, exact token match, edit-distance
(Levenshtein) fuzzy similarity, and a phonetic key, with weights
$(w_e,w_f,w_p)=(1.0,1.2,0.6)$. An exact token hit is the most reliable signal,
fuzzy similarity is up-weighted slightly per unit because it is a continuous
score in $[0,1]$ {and} is the
primary mechanism for recovering near-miss entities. The phonetic cue is binary
and coarse, so it is down-weighted. Normalized Levenshtein similarity is
$\text{sim}(a,b)=1-\text{lev}(a,b)/\max(|a|,|b|)$, where words are compared only when
their lengths differ by $\le 3$ characters and have $\ge 4$ characters (a
speed or quality filter), and phonetic keys use a Double-Metaphone-style
5-character prefix. The score only \emph{ranks} candidates and we take the
top-$K$ ($K=200$). The weights and $K$ are ablated in
Table~\ref{tab:retrieval}, which shows $K=200$ is best and that combining all
three signals (with fuzzy up-weighted) beats any single signal or uniform
weights.
\paragraph{Candidate budget $K$.}
$K_{\text{init}}=\min(200,|\mathcal{E}|)$ with a feedback ceiling
$K_{\text{cap}}=\min(400,|\mathcal{E}|)$. The $\min(\cdot,|\mathcal{E}|)$ sends
small entity lists in full. As Table~\ref{tab:retrieval} shows, \ewer{} is
minimized at $K=200$ and degrades slightly for $K=400$ and the full list, since
beyond a few hundred candidates the added phrases are low-scoring and dilute the
prompt without improving recall.

\paragraph{Larger $K$ during feedback.}
\label{app:kfeedback}
The ablation in Table~\ref{tab:retrieval} holds $K$ fixed on feedback to isolate
the effect of the budget. Our main runs instead raise $K$ from $200$ to
$K_{\text{cap}}=400$ when a feedback round fires. Table~\ref{tab:kfeedback}
compares the two on the same setting (Earnings-21, Whisper-large-v3,
GPT-4o-mini, LLM-Select). A larger static budget is worse ($11.64$ at $K=400$
against $11.46$ at $K=200$), but raising $K$ only during feedback improves
\ewer{} to $11.40$. Increasing the budget globally adds noise, while targeting a
wider pool at the span the critic flagged recovers a few more entities.

\begin{table}[h]
  \centering
  \caption{Effect of the retrieval budget during feedback (Earnings-21,
  Whisper-large-v3, GPT-4o-mini, LLM-Select). \ewer\,\% with RWERR vs.\ baseline
  in parentheses.}
  \label{tab:kfeedback}
  \small
  \begin{tabular}{@{}lc@{}}
    \toprule
    \textbf{Retrieval budget} & \textbf{\ewer\,$\downarrow$ (RWERR)} \\
    \midrule
    $K=200$, fixed on feedback & 11.46 (27.8) \\
    $K=400$, fixed on feedback & 11.64 (26.7) \\
    $K=200 \to 400$ on feedback & \textbf{11.40 (28.2)} \\
    \bottomrule
  \end{tabular}
\end{table}

\paragraph{Guardrail thresholds.}
The \textsc{Verify} similarity floor is $0.5$ and it blocks
unrelated substitutions (\textit{star}$\rightarrow$\textit{cytiva}) while
admitting near-misses (\textit{citeva}$\rightarrow$\textit{cytiva}). These are coarse safety bounds,
not tuned values. The critic, not the threshold, is the precision decider.

\paragraph{Feedback.}
Feedback is bounded to $R=1$ round in our experiments and is {evidence-gated}: a
critic-flagged missed span triggers a round only if a list entity matches it at
similarity $\ge\tau_a\!=\!0.6$ and differs from the current text (otherwise the
loop stops). Over-corrections are dropped. Off-list proposals are snapped, not
re-run. The critic uses the same LLM as correction here.

\section{Extended Scoring-weight Grid}
\label{app:weightgrid}
Table~\ref{tab:weightgrid} reports a finer grid around the reported weights, varying one weight at a time (Earnings-21, Whisper-large-v3, GPT-4o-mini). \ewer{} spans $11.40$--$11.69$ across the grid, so the reported setting is on a plateau and it is no worse than any neighbour. The coarser sweep in Table~\ref{tab:retrieval}, which switches signals off entirely, shows much larger movement (up to $13.93$), so what matters is that all three signals are present and that fuzzy similarity is not down-weighted.

\begin{table}[h]
  \centering
  \caption{Scoring-weight grid at $K\!=\!200$ (Earnings-21, Whisper-large-v3, GPT-4o-mini, LLM-Select).}
  \label{tab:weightgrid}
  \small
  \setlength{\tabcolsep}{6pt}
  \begin{tabular}{@{}lc@{}}
    \toprule
    \textbf{Weights $(w_e,w_f,w_p)$} & \textbf{\ewer\,$\downarrow$} \\
    \midrule
    $(1.0,\ 1.2,\ 0.6)$ reported & \textbf{11.40} \\
    $(1.0,\ 1.5,\ 0.6)$ & \textbf{11.40} \\
    $(1.0,\ 1.2,\ 0.3)$ & 11.41 \\
    $(1.5,\ 1.2,\ 0.6)$ & 11.47 \\
    $(1.0,\ 0.8,\ 0.6)$ & 11.69 \\
    \bottomrule
  \end{tabular}
\end{table}

\section{Hypothesis Diversity}
\label{app:diversity}
Table~\ref{tab:diversity} quantifies the evidence that the alternative hypotheses carry, holding the Whisper-large-v3 1-Best fixed so that only the alternatives change. \emph{Entity evidence only in alternatives} is the fraction of reference entity tokens that appear correctly in at least one alternative hypothesis but not in the 1-Best, that is, the evidence a 1-Best-only corrector cannot see. \emph{Segments carrying such evidence} is the fraction of segments containing at least one such token. \emph{Entity coverage} is the fraction of reference entity tokens present in at least one supplied hypothesis, and the column reports its increase from $N\!=\!1$ to $N\!=\!5$. Independent recognizers supply roughly three to four times the recoverable evidence of temperature sampling, which is the ceiling any fusion strategy operates under and explains the gap between Table~\ref{tab:main} and Table~\ref{tab:crossasr}.

\begin{table}[h]
  \centering
  \caption{Entity evidence available in the alternative hypotheses, Earnings-21, with the Whisper-large-v3 1-Best held fixed.}
  \label{tab:diversity}
  \small
  \setlength{\tabcolsep}{4pt}
  \resizebox{\columnwidth}{!}{%
  \begin{tabular}{@{}lcc@{}}
    \toprule
    & \textbf{\makecell{Temperature\\sampling}} & \textbf{\makecell{Five\\recognizers}} \\
    \midrule
    Entity evidence only in alternatives & 4.90\% & \textbf{16.94\%} \\
    Segments carrying such evidence & 6.28\% & \textbf{21.00\%} \\
    Entity coverage, $N{=}1$ to $N{=}5$ & $+$5.15\% & \textbf{$+$20.40\%} \\
    \bottomrule
  \end{tabular}}
\end{table}

\section{Entity-list Coverage and Noise}
\label{app:listcov}
This section details how the degraded and noisy lists of Table~\ref{tab:listcov} were built.

\paragraph{List construction.} The starting point is the 1{,}013-phrase oracle entity list released with Earnings-21~\cite{del2021earnings}. The degraded lists retain the stated fraction of entities under a single fixed random seed, and removal targets entities that actually occur in the references, so that each condition is a genuine loss of recoverable corrections rather than the removal of phrases the retriever would never have surfaced. The noisy condition uses the published distractor list of~\citet{fox2022improving} as released: it keeps all 1{,}013 oracle phrases and adds 769 in-domain company and executive names that do \emph{not} occur in Earnings-21, giving 1{,}782 entries. In-domain confusable distractors are a harder test than random out-of-domain strings, which the retriever would rank below any hypothesis-matching phrase. Retrieval, prompts, guardrail and critic are unchanged across all conditions. \ewer{} is always computed against the full reference entity set, so a phrase removed from the list still counts against the system when it is missed.
\section{Cost and Latency}
\label{app:cost}
Table~\ref{tab:cost} in Section~\ref{sec:cost} gives cost and latency for the main configurations. Two further measurements, taken the same way: RECOVER \emph{without} the feedback round on GPT-4o uses 2.00 calls and 3{,}979\,$+$\,327 tokens per segment at USD 12.94 per 1{,}000 segments, so the feedback round accounts for 0.11 calls and about 5\% of cost, and the single-pass entity-list corrector costs USD 5.01 per 1{,}000 segments on GPT-4o (1{,}222\,$+$\,196 tokens, 3.66\,s) against USD 0.29 on GPT-4o-mini.
\section{Open-weight Correctors}
\label{app:openweight}
Table~\ref{tab:openweight} runs the RECOVER pipeline, retriever, and prompts unchanged with two open-weight models as both corrector and critic, served through OpenRouter~\cite{openrouter}. Llama-3.3-70B sits between GPT-4o-mini and GPT-4o on accuracy and Mixtral-8x22B is level with GPT-4o-mini, so the framework does not depend on a proprietary model, although the strongest result still comes from the strongest LLM.

\begin{table}[h]
  \centering
  \caption{Corrector and critic LLM on Earnings-21 (Whisper-large-v3, LLM-Select), with RECOVER otherwise unchanged. Cost and latency are measured as in Table~\ref{tab:cost}.}
  \label{tab:openweight}
  \scriptsize
  \setlength{\tabcolsep}{3pt}
  \resizebox{\columnwidth}{!}{%
  \begin{tabular}{@{}lcccccc@{}}
    \toprule
    \textbf{LLM} & \textbf{\wer} & \textbf{\ewer} & \textbf{RWERR} & \textbf{calls} & \textbf{USD/1k} & \textbf{med.\ s} \\
    \midrule
    No correction (baseline) & 12.34 & 15.88 & -- & -- & -- & -- \\
    \midrule
    GPT-4o & 12.39 & \textbf{9.76} & \textbf{38.5} & 2.11 & 13.57 & 6.14 \\
    GPT-4o-mini & 12.48 & 11.40 & 28.2 & 2.35 & 0.84 & 5.57 \\
    \midrule
    Llama-3.3-70B & 12.50 & 11.12 & 30.0 & 2.20 & \textbf{0.60} & 10.89 \\
    Mixtral-8x22B & 12.51 & 11.42 & 28.1 & 2.34 & 12.51 & \textbf{4.32} \\
    \bottomrule
  \end{tabular}}
\end{table}

Costs for the open-weight models use OpenRouter list prices at the time of writing (Llama-3.3-70B USD 0.10 / USD 0.32 and Mixtral-8x22B USD 2.00 / USD 6.00 per million prompt / completion tokens) and are hosted-inference prices, not the cost of self-hosting. Read together with accuracy, Llama-3.3-70B is the best cost-for-quality point we measured: cheaper than GPT-4o-mini (USD 0.60 against USD 0.84 per 1{,}000 segments) and better on entities (11.12 against 11.40 \ewer{}), at roughly twice the latency (10.89\,s against 5.57\,s), which is a reasonable trade in a batch setting and a poor one in a streaming setting. Mixtral is the fastest configuration but at current hosted prices costs almost as much as GPT-4o for a worse result. Both open-weight models emit close to twice the completion tokens of the GPT models (529--540 against 277--333), which is where their latency goes. For reference on the single-pass side, the entity-list prompt of Table~\ref{tab:cost} costs USD 5.01 per 1{,}000 segments on GPT-4o (1{,}222 $+$ 196 tokens, 3.66\,s) against USD 0.29 on GPT-4o-mini. Tokens per segment are 4{,}283 $+$ 540 for Llama-3.3-70B and 4{,}669 $+$ 529 for Mixtral-8x22B.

\section{Adaptive Routing Analysis}
\label{app:routing}
The Adaptive Routing configuration uses the LLM-as-router to pick one fusion strategy per segment from the
inter-hypothesis agreement and the per-variant entity-hit counts.
Table~\ref{tab:routing} reports how often each strategy is selected across all
segments of each dataset (Whisper-large-v3, GPT-4o). The router overwhelmingly
chooses Entity-Aware Select (60--97\% of segments) and selects LLM-Select on
under 8\% of segments, even though LLM-Select is the strongest fixed strategy in
Table~\ref{tab:main}. This explains why adaptive routing does not improve over a
fixed LLM-Select. This likely needs richer routing signals or a
trained router, which we leave to future work.

\begin{table}[h]
  \centering
  \caption{Adaptive Routing: number of segments assigned to each fusion strategy
  by the LLM router (Whisper-large-v3, GPT-4o).}
  \label{tab:routing}
  \small
  \setlength{\tabcolsep}{4pt}
  \resizebox{\columnwidth}{!}{%
  \begin{tabular}{@{}lcccc@{}}
    \toprule
    \textbf{Dataset} & \textbf{\makecell{Entity-Aware\\Select}} & \textbf{\makecell{ROVER\\Ensemble}} & \textbf{\makecell{LLM-\\Select}} & \textbf{1-Best} \\
    \midrule
    Earnings-21 & 1{,}745 & 252 & 89 & 0 \\
    ATCO2 & 337 & 172 & 43 & 8 \\
    Eka-Medical & 2{,}953 & 637 & 29 & 0 \\
    Common Voice & 15{,}909 & 426 & 66 & 0 \\
    ContextASR-Bench & 4{,}524 & 369 & 380 & 0 \\
    \bottomrule
  \end{tabular}}
\end{table}

\begin{table}[h]
  \centering
  \caption{Effect of disabling the critic and feedback on Earnings-21
  (Whisper-large-v3, GPT-4o, LLM-Select). All values in \%.}
  \label{tab:criticwer}
  \small
  \setlength{\tabcolsep}{4pt}
  \resizebox{\columnwidth}{!}{%
  \begin{tabular}{@{}lcccccc@{}}
    \toprule
    \textbf{System} & \textbf{\wer} & \textbf{\ewer} & \textbf{RWERR} & \textbf{P} & \textbf{R} & \textbf{F1} \\
    \midrule
    Baseline & 12.34 & 15.88 & -- & 92.73 & 78.95 & 85.29 \\
    LLM-Select (w/ critic+fb) & 12.39 & 9.76 & 38.5 & 89.82 & 87.36 & 88.57 \\
    \;\;w/o critic+fb & 12.40 & 10.2 & 36.0 & 89.6 & 87.0 & 88.3 \\
    \bottomrule
  \end{tabular}}
\end{table}

\section{Effect of the Critic and Feedback on Overall WER}
\label{app:criticwer}
The feedback fires on few segments and the critic rejects only a small number
of edits with
the critic and feedback disabled, keeping only the deterministic guardrail. On Earnings-21 (Whisper-large-v3, GPT-4o,
LLM-Select) the critic rejects 112 edits and feedback adds a correction on about
1.2\% of the 2{,}086 segments (i.e., 25 segments). The overall \wer{} changes by $+0.01$\,pp (12.39 to 12.40), which is
negligible and still close to the 12.34 baseline. The change is a little more visible, with \ewer{} rising from
9.76 to about 10.2 (RWERR 38.5 to about 36) and F1 falling from 88.57 to about
88.3 (Table~\ref{tab:criticwer}). The critic and feedback therefore mainly
improve entity precision and recall rather than overall transcript quality.

\end{document}